\documentclass[10pt,conference,a4paper]{IEEEtran}

\usepackage{amssymb}
\usepackage{bm}
\usepackage{amsmath,graphicx}
\usepackage{tabularx}
\usepackage[ruled,norelsize]{algorithm2e}
\usepackage[dvipsnames]{xcolor}

\begin{document}
\title{Space-Time Domain Tensor Neural Networks: \\An Application on Human Pose Classification}

%\author{\IEEEauthorblockN{Konstantinos Makantasis}
%\IEEEauthorblockA{Institute of Digital Games\\
%University of Malta\\
%konstantinos.makantasis@um.edu.mt}}
%\and
%\IEEEauthorblockN{Homer Simpson}
%\IEEEauthorblockA{Twentieth Century Fox\\
%\and
%\IEEEauthorblockN{James Kirk\\ and Montgomery Scott}
%\IEEEauthorblockA{Starfleet Academy\\
%San Francisco, California 96678--2391\\
%elephone: (800) 555--1212\\


\author{\IEEEauthorblockN{Konstantinos Makantasis\IEEEauthorrefmark{1},
Athanasios Voulodimos\IEEEauthorrefmark{2},
Anastasios Doulamis\IEEEauthorrefmark{3},\\
Nikolaos Bakalos\IEEEauthorrefmark{3} and
Nikolaos Doulamis\IEEEauthorrefmark{3}}
\IEEEauthorblockA{\IEEEauthorrefmark{1}Institute of Digital Games, University of Malta}
\IEEEauthorblockA{\IEEEauthorrefmark{2}Department of Informatics and Computer Engineering, University of West Attica}
\IEEEauthorblockA{\IEEEauthorrefmark{3}School of Rural and Surveying Engineering, National Technical University of Athens}}

\maketitle

\begin{abstract}
Recent advances in sensing technologies require the design and development of pattern recognition models capable of processing spatiotemporal data efficiently. In this study, we propose a spatially and temporally aware tensor-based neural network for human pose classifiaction using three-dimensional skeleton data. Our model employs three novel components. First, an input layer capable of constructing highly discriminative spatiotemporal features. Second, a tensor fusion operation that produces compact yet rich representations of the data, and third, a tensor-based neural network that processes data representations in their original tensor form. Our model is end-to-end trainable and characterized by a small number of trainable parameters making it suitable for problems where the annotated data is limited. Experimental evaluation of the proposed model indicates that it can achieve state-of-the-art performance. %Our methodology is general enough and can potentially be applied to different pattern recognition problems employing spatiotemporal data from sensor networks. In this study, however, we consider  the problem of human pose classification to demonstrate the efficiency of our methodology.
\end{abstract}

% \fi
\IEEEpeerreviewmaketitle

\section{Introduction}
\label{sec:intro}
Advances in sensing technologies have enabled the development of time-evolving sensor networks where a single node can monitor a plethora of user (e.g. body sensor networks) and environmental information \cite{gravina2017multi}. Sensed information corresponds to multimodal data, in space and time, which is used to continuously observe the progress of a phenomenon \cite{vuran2004spatio}. Processing and correlating multiple, potentially heterogeneous, information streams to detect and recognize spatiotemporal patterns is becoming a fundamental yet non-trivial task. %A typical and emerging example of spatiotemporal sensing is Kinect-II 3D skeleton information, which extracts, in case of humans, 3D point joints and their motion in space and time.     

The presented work focuses on processing and fusing data coming from multiple information streams, as well as on discovering informative patterns for a given learning task at hand. Specifically, we introduce a novel tensor-based deep neural network model able to automatically process and correlate spatiotemporal information from different sources and discover appropriate patterns for assigning inputs to desired outputs. This is a generic space-time learning machine, which can be useful for a variety of time series analysis applications, such as human's behavior recognition, moving objects analysis, radar signals, audio processing, etc. In this study, however, we consider the problem of human pose classification using 3D skeleton information coming from the Kinect-II sensor \cite{zhang2012microsoft} to demonstrate the efficiency of our methodology.

Initially, the tensor-based neural network processes 3D skeleton information to extract spatiotemporal patterns that can collectively describe specific human poses. Then, the derived patterns are fused into a rich yet compact tensor object, which, in turn, is processed by the tensor-based neural network. Although the steps mentioned above seem to be independent, actually they happen concurrently within an \textit{end-to-end trainable} tensor-based learning machine.

The input layer of the proposed model is designed to process spatiotemporal data for constructing compact and discriminative features for the learning task at hand. The design of the input layer is inspired by the Common Spatial Patterns (CSP) algorithm \cite{grosse2008multiclass}, and thus, we refer to the constructed features as CSP-like features. The constructed features for each information stream are fused into a compact yet rich tensor representation, which in turn is processed by sequential tensor contraction layers. The number of tensor construction layers determines the depth of the learning machine, enabling, this way, the design of deep tensor-based learning architectures. 

\subsection{Related Work}
This paper deals with spatiotemporal information streams and their processing, and thus, this section is divided into three subsections; correlating multiple information streams, pattern analysis for spatiotemporal data, and human pose classification.

\subsubsection{Correlating Multiple Information Streams}
\label{sssec:fusion}
Fusion techniques merge and correlate information from different data streams. These techniques can be classified into feature-level and score-level. Score-level fusion methods select a hypothesis based on a set of hypotheses generated by processing each data stream \textit{separately}  \cite{wolpert1992stacked, simonyan2014two}. The final hypothesis is selected either by averaging the generated hypotheses or by stacking another learning machine. In the latter case, the input of the learning machine is the set of generated hypotheses from each stream and its output the final hypothesis. Score-level fusion approaches do not correlate the information from different data streams; instead, they try to make robust decisions by operating similar to ensemble methods. Feature-level fusion approaches aggregate features or raw data from different data streams by element-wise averaging or addition (assuming that the dimension of features allows it) or by concatenation \cite{park2016combining, liapis2019fusing}. However, these simple feature aggregation techniques cannot capture complex interactions between different data streams. Therefore, capturing and modeling such interactions is left to a learning machine that follows the fusion operation. 

Although learning machines are capable of disentangling complex relations in data \cite{lecun2015deep}, fusion techniques capable of highlighting such relations \cite{makantasis2019common} are crucial for the successful training, especially under small sample setting problems that employ a limited number of training examples.  The work presented in \cite{hu2017attribute} tries to overcome the above limitation by proposing a rich tensor-based data fusion framework. Various data streams are fused into a unified tensor object whose dimension, then, is reduced via Tucker decomposition. 

In this study, we fuse 3D skeleton information into unified multilinear (tensor) objects following the approach in \cite{hu2017attribute}. We do not, however, decompose the fused information to create appropriate inputs (e.g. matrices or vectors) for the employed learning machine. Instead, we use a tensor-based learning machine capable of processing the fused information in its original multilinear form.  

\subsubsection{Pattern Recognition for Spatiotemporal Data}
\label{sss:recognition}
Efficient pattern recognition algorithms for processing spatiotemporal information aim to discover and correlate patterns across both the spatial and the temporal domain of the data. The discovery of such patterns is related to the feature construction process, while the correlation of those patterns to the employment of a machine learning model. Those two processes can be conducted separately or fused into a unified framework. In the first case, features %, which are compact representations of the spatiotemporal information of the data,
are constructed and then are used as input to machine learning models. In the latter case, the feature construction process takes place during the training of models by using, for example, deep learning architectures.

The most common approach for compactly representing spatiotemporal data is by using statistical features such as mean, variance, energy and entropy %\cite{bao2004activity, 
\cite{wang2005human, ravi2005activity, makantasis2017data}. By treating spatiotemporal data as time series, frequency domain features, such as Fast Fourier \cite{huynh2005analyzing} and Wavelets transform \cite{abdu2018novel} coefficients, can also be used to represent the data. More sophisticated approaches employ Autoregressive models \cite{he2008activity, khan2008accelerometer} to construct features for representing spatiotemporal data via a learning (model-fitting) process. The approaches mentioned above focus solely on feature construction. Therefore, there is no information flow between the feature construction and the pattern recognition tasks, even though these are sequential. That poses several problems, such as computational complexity, difficulty in transfer of learning and adaptation, and, in many cases, a high-risk for over-fitting \cite{nikitakis2019unified}.

Deep learning models unify the feature construction and pattern recognition tasks by learning high-level representations of raw inputs during the training phase. Convolutional Neural Networks (CNNs) are state-of-the-art learning machines for processing spatial data. Besides spatial data, CNNs can also process spatiotemporal data, either directly by using spatiotemporal convolutions \cite{tran2015learning, ji20123d}, or indirectly by applying spatial convolutions on spatiotemporal data \cite{karpathy2014large, makantasis2016deep}, for example on videos where frames are concatenated along the temporal dimension. %When the data are spatially coherent, i.e., neighbouring bits of information are highly correlated (e.g., pixels in images), then CNNs can produce highly descriptive features. When, however, such coherency is not the case (e.g., EEG data where the responses of adjacent channels/electrodes are not necessarily related), CNNs are not able to produce high-quality features. Besides the requirement for spatially coherent data, another 
The most important drawback, however, of deep learning models is the number of their trainable parameters. Usually, those models employ a vast number of parameters (much larger than the number of available data) the values of which is tough to be estimated especially when small sample setting problems need to be addressed \cite{makantasis2018tensor}. 

In this work, we propose a machine learning model capable of unifying the feature construction and pattern recognition tasks, and at the same time, it overcomes the main limitation of deep learning models. Specifically, by exploiting tensor algebra tools, we can significantly reduce the number of model’s trainable parameters making it suitable for problems where the number of available data is limited. %Second, the proposed model can capture spatial correlations even for data that are not spatially coherent by employing a novel neural network layer capable of constructing CSP-like features. The design of this layer is inspired by the CSP algorithm, which does not require spatial coherency within the data. 

\subsubsection{Human Pose Classification}
Human pose classification is usually formulated as a computer vision problem, where the human poses are described via the detection of body parts through pictorial structures \cite{toshev2014human, chen2014articulated, tompson2014joint}. In this study, instead of using visual information, we focus on human pose classification using solely 3D skeleton measurements. 3D skeleton data are used in \cite{raptis2011real} for the development of a gesture classification system. The authors of \cite{zanfir2013moving} propose the Moving Pose system, which is based on a 3D kinematics descriptor. In \cite{kitsikidis2014dance}, skeleton data is split into different body parts, which are then transformed to allow view-invariant pose recognition. 3D skeleton data from MS Kinect are used in \cite{ball2012unsupervised} for recognizing individual persons based on their walking gait, while Rallis \textit{et al.} in \cite{rallis2017extraction} propose a key posture identification method based on Kinect-II measurements. In \cite{guerra2020automatic} 3D skeleton data from Kinect One are used to develop an automatic pose recognition system for assisting disabled students living in college dorms. Finally, the study in \cite{wang2019comparative} presents a recent comparative review of action recognition algorithms that use Kinect data.

\subsection{Our Contribution}
Based on the discussion so far, the main contributions this study can be summarized into the following three points. First, we propose an \textit{end-to-end} trainable architecture that unifies the feature and pattern recognition tasks. Second, we exploit tensor algebra tools to significantly reduce the number of the proposed model's trainable parameters making it very robust for small sample setting problems. Last but not least, the proposed approach is a \textit{general} one that can potentially be applied to different problems that employ spatiotemporal data coming from sensor networks. In this study, however, we focus on the problem of human pose classification for demonstrating the efficiency of our proposed model.

\section{Approach Overview}
In this section, we formulate the problem that we are trying to address and present the main components of the proposed methodology. For the rest of the paper, we represent scalars, vectors, matrices and tensor objects of order larger than two with lowercase, bold lowercase, uppercase and bold uppercase letters respectively. 

\subsection{Problem Formulation}
\label{ssec:formulation}
We consider the problem of human pose classification using 3D skeleton data from Kinect-II. As we will see later, that problem is a specific instance of the more general problem of pattern recognition using information coming from sensor networks. Therefore, in this section, we describe the form of the latter more general problem. 

Consider a sensor network that contains $C$ sensors. Each one of the sensors, let's say the $c$-th sensor, retrieves $J$ measurements (information modalities) at each time instance $t$, which can be represented by the vector
\begin{equation}
\label{eq:sensor}
\bm s_c(t) = [x_c^{(1)}(t), \cdots , x_c^{(j)}(t), \cdots , x_c^{(J)}(t)]
\end{equation}
for $c=1,\cdots, C$. Since each sensor occupies a specific spatial position, the spatial information for the $j$-th information modality captured by the sensor network can be represented by the following vector:
\begin{equation}
\bm s^{(j)}(t) = [x_1^{(j)}(t), x_2^{(j)}(t), \cdots, x_C^{(j)}(t)]
\end{equation}
for $=1,\cdots, J$, while the spatiotemporal information corresponding to a time window $t$ to $t+T$ can be represented by the matrix
\begin{equation}
\label{eq:modality}
S^{(j)}(t, t+T) = [\bm s^{(j)}(t); \cdots ; \bm s^{(j)}(t+T)]^\top \in \mathbb R^{C \times T}.
\end{equation}
The information from all $S^{(j)}(t, t+T)$, $j=1,\cdots, J$ can be aggregated into a tensor object
\begin{equation}
\label{eq:tensor_object}
\bm S(t, t+T) = [S^{(1)}(t, t+T); \cdots S^{(J)}(t, t+T)]
\end{equation}
in $\mathbb R^{C \times T \times J}$. For the sake of clarity, in the following we omit the time index, thus, when we write $\bm S$ we refer to a tensor object of the form of (\ref{eq:tensor_object}) for some time instance $t$. Obviously, for a specific time window, the tensor object in (\ref{eq:tensor_object}) encodes the spatiotemporal information for all information modalities and all sensors in a sensor network. 

Each tensor $\bm S$ describes a pattern that belongs to a specific class. Let us denote as $y$ the class of that pattern, and assume that we have in our disposal a set $\mathcal D$ of $N$ pairs of the form:
\begin{equation}
\label{eq:training_set}
\mathcal D = \big\{(\bm S_i, y_i) \big\}_{i=1}^N.
\end{equation} 
The objective of this study is to derive a function for mapping $\bm S$ to $y$ given the set $\mathcal D$ in (\ref{eq:training_set}). This can be seen as a machine learning problem. Let us denote as $\mathcal F$ the class of functions that can be computed by a learning machine. We want to select the function 
\begin{equation}
\label{eq:problem}
f^* =\arg \min_{f\in \mathcal F} \sum_i l(f(\bm S_i), y_i)
\end{equation}
such that $(\bm S_i, y_i) \in \mathcal D$. In (\ref{eq:problem}) $l(\cdot)$ is a loss function. For classification problems $l(\cdot)$ usually is the cross entropy loss.

\noindent \textit{Remark 1}: In order to facilitate the solution of problem (\ref{eq:problem}) the learning machine must contain a number of trainable parameters that is comparable to the cardinality $N$ of set $\mathcal D$, and at the same time it should be capable of fully exploiting the spatiotemporal nature of the data. 

\noindent \textit{Remark 2}: The problem of human pose recognition using 3D skeleton data from Kinect-II is a special instance of the problem described above. Each skeleton joint can be seen as a sensor, which, at every time instance, measures its $x-y-z$ location. So, in this case $C$ equals the number of skeleton joints and $J$ in (\ref{eq:sensor}) equals 3 ($x$, $y$ and $z$ positions).
 
\subsection{Proposed Methodology}
In this study, we use 3D skeleton data captured using Kinect-II, along with their annotations, which correspond to the depicted human pose at every time instance. Initially, we process the skeleton data to create tensor objects as in (\ref{eq:tensor_object}) and then use their annotations to create a training set as in (\ref{eq:training_set}). 

After creating the training set, we design an \textit{end-to-end} trainable neural network, which is able to fully exploit the spatiotemporal nature of the data, and at the same time employs a small number of trainable parameters (compared to the size of the training set). The first layer of the proposed model learns \textit{CSP-like} features \cite{nikitakis2019unified} from each information modality using inputs in the form of (\ref{eq:modality}). Then, the constructed features from all modalities are fused into a tensor object to compactly represent the spatiotemporal information captured by the sensor network. Finally, the tensor objects are processed by a tensor-based neural network for producing a mapping from 3D skeleton data to human poses. In the following, we describe each one of the steps presented above in details.

\begin{figure}[t]
	\begin{minipage}{\linewidth}
		\centering
		\centerline{\fbox{\includegraphics[width=1.0\linewidth]{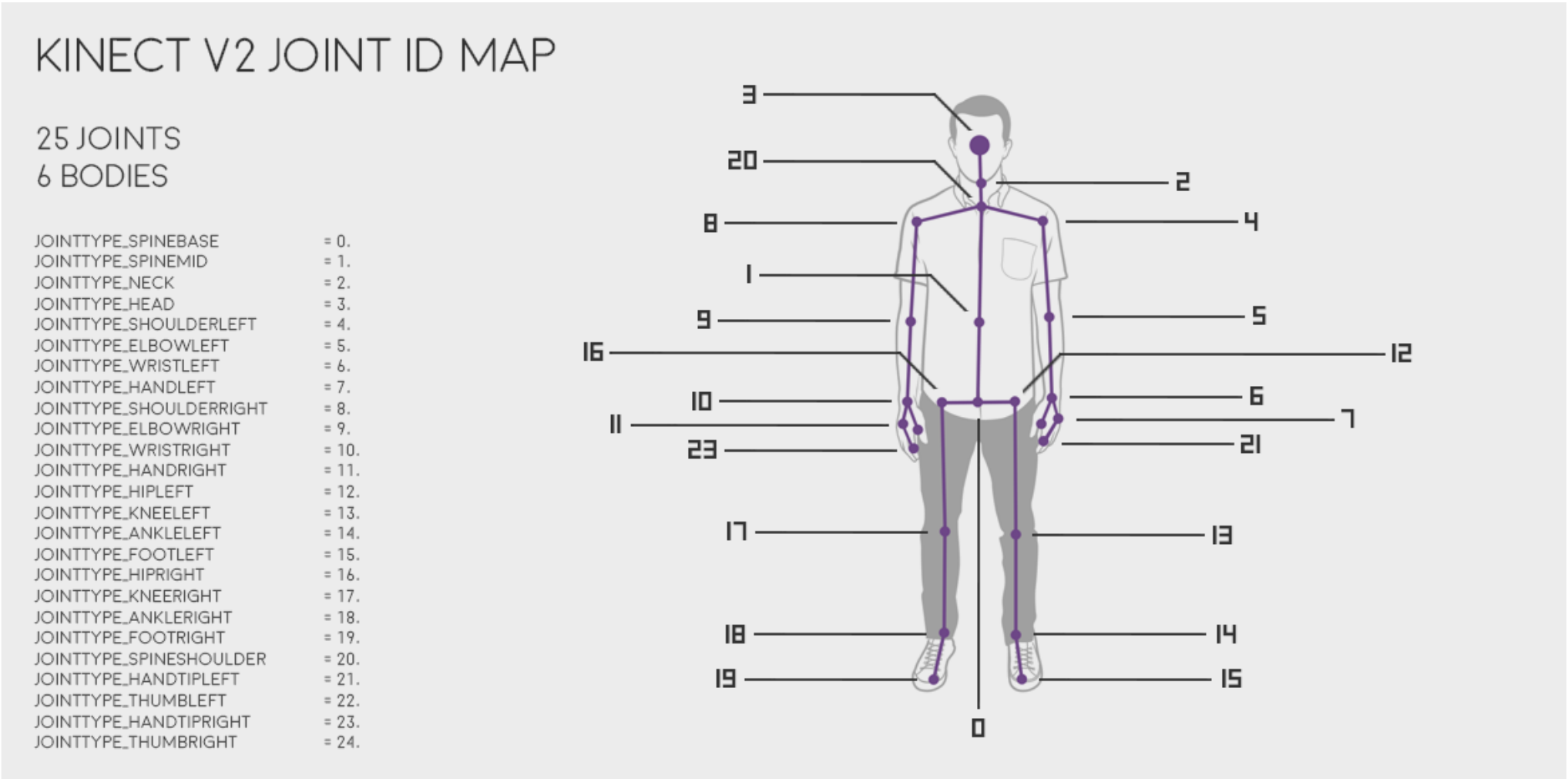}}}
	\end{minipage} 
	\caption{Kinect II skeletal capturing system (vvvv.org/documentation/kinect).}
	\label{fig:kinect}
\end{figure}

\section{Data Preprocessing}
\label{sec:preprocessing}
%In this section, we present the 3D skeleton data as well as their preprocessing for human pose classification. 
The Kinect-II sensor identifies and monitors twenty-five skeletal joints at the constant rate of 30 measurements per second, see Fig. \ref{fig:kinect}. The positions of joints in the 3D space with respect to the Kinect-II device are provided. We utilise the measurements in the form they are captured without employing any tracking technique. A human pose, however, is characterized by the relative positions of the human body parts. For this reason, we represent the position of each joint with respect to the position of the Spine Base joint. %In other words, we use the Spine Base joint as the origin of a local coordinate system. 
This way, the recognition of human poses does not depend on the position of the human with respect to the Kinect-II device.

Specifically, if we denote as $\bm s'_0(t)$ the coordinates of the Spine Base joint and as $\bm s'_c(t)$, $c=1,\cdots,24$ the coordinates of all other joints, then the coordinates of the joints with respect to the Spine Base joint will be given by 
\begin{equation}
\label{eq:local_coord}
\bm s_c(t) = \bm s'_c(t) - \bm s'_0(t) \:\:,\:\: c=1,\cdots, 24.
\end{equation}
Using the transformed coordinates in (\ref{eq:local_coord}), we create matrices as in (\ref{eq:modality}) for $j=1,\cdots,3$ that correspond to $x-y-z$ positions. Those matrices encode the spatiotemporal information for classifying human poses.%, and, thus, we want to map those matrices to a specific human pose. 

At this point, we have to mention that parameter $T$ in (\ref{eq:modality}) is application dependent and affects the recognition results. For this reason, it must be set appropriately. For $T=1$, the pose recognition model will not be able to exploit the temporal information and thus it will be more prone to measurements errors, while large values of $T$ may result to a dataset where each datum depicts more than one pose, increasing, this way, the uncertainty in recognition. The effect of parameter $T$ on the recognition results is further discussed in Section \ref{sssec:parameterT}.  

\section{Space-Time Domain Tensor Neural Network}
The proposed tensor-based neural network consists of three main components; the input layer capable of computing CSP-like features, the tensor fusion operation, and the tensor contraction and regression layers that process high-order data in its original multilinear form. %In the following, we describe each one of those components in detail.

\subsection{CSP Neural Network Layer}
The CSP layer aims to produce highly discriminative features for human pose classification. The design of that layer is motivated by the CSP algorithm \cite{grosse2008multiclass}, which, for the sake of clarity and completeness, we briefly describe here.%is widely used for classifying EEG signals. For the sake of clarity and completeness, we briefly describe the CSP algorithm. 

The CSP algorithm originally was developed for binary classification problems. It receives as input zero average signals in the form of (\ref{eq:modality}) along with their labels. Then, its objective is to produce features that increase the separability between two pattern classes. Consider that we have in our disposal $N$ samples $\{S_{l,i}\}_{i=1}^N$, where $l=1,2$ denotes the class of each sample. The CSP algorithm computes the covariance matrix
\begin{equation}
R_{l,i} =\frac{S_{l,i} S_{l,i}^\top}{\text{trace}(S_{l,i} S_{l,i}^\top)}
\end{equation}
for each sample, and the average covariance matrix
\begin{equation}
\bar{R}_l = \frac{1}{n_l} \sum_{i=1}^{n_l} R_{l,i} , \:\: l=1,2, 
\end{equation}
for each class, where $n_l$ is the number of samples belonging to class $l$. Then, the CSP filter, $W$, is constructed by using $M=2m$, $(M < C)$, eigenvectors corresponding to $m$ largest and $m$ smallest eigenvalues of $\bar{R}_2^{-1}\bar{R}_1$. Finally, using $W$ each sample is represented by a feature vector %$\bm f_{l,i}\in \mathbb R^M$ of the following form;
\begin{equation}
\label{eq:csp_features}
\bm f_{l,i} = \log \Bigg[\frac{\text{var} (Y_{l,i}^1)}{\sum_{j=1}^M \text{var} (Y_{l,i}^j)} \cdots \frac{\text{var} (Y_{l,i}^M)}{\sum_{j=1}^M \text{var} (Y_{l,i}^j)}\Bigg] \in \mathbb R^M,
\end{equation}
where $Y_{l,i}^j$ stands for the $j$-th row of $WS_{l,i}$. Features $\bm f_{l,i}$ typically are used for as inputs to learning models since they encode the spatiotemporal information of signals $\{S_{l,i}\}_{i=1}^N$. 

Although, theoretically sound, the CSP algorithm presents several drawbacks when applied to real world problems mainly due to the non-stationarity of captured signals \cite{nikitakis2019unified}. Moreover, it is a feature construction technique that is performed individually, and thus does not permit information flow between feature construction and pattern recognition tasks (see Section \ref{sss:recognition}). 
To overcome those drawbacks, the proposed CSP layer learns $W$ during the training of phase model  \cite{nikitakis2019unified}. Trainable matrix $W$ projects measurements in $\mathbb R^{M \times T}$ and then features as in (\ref{eq:csp_features}) are computed from the projected measurements. Additionally, since Kinect-II measurements extract 3D coordinates, we use three parallel CSP layers, one for each coordinate. Therefore, the output of the CSP layer consists of three vectors in $\mathbb R^M$.  %Finally, as shown in \cite{nikitakis2019unified} parameters $W$ can be efficiently learned in an end-to-end trainable neural network.  

\begin{figure}[t]
	\begin{minipage}{\linewidth}
		\centering
		\centerline{\fbox{\includegraphics[width=1.0\linewidth]{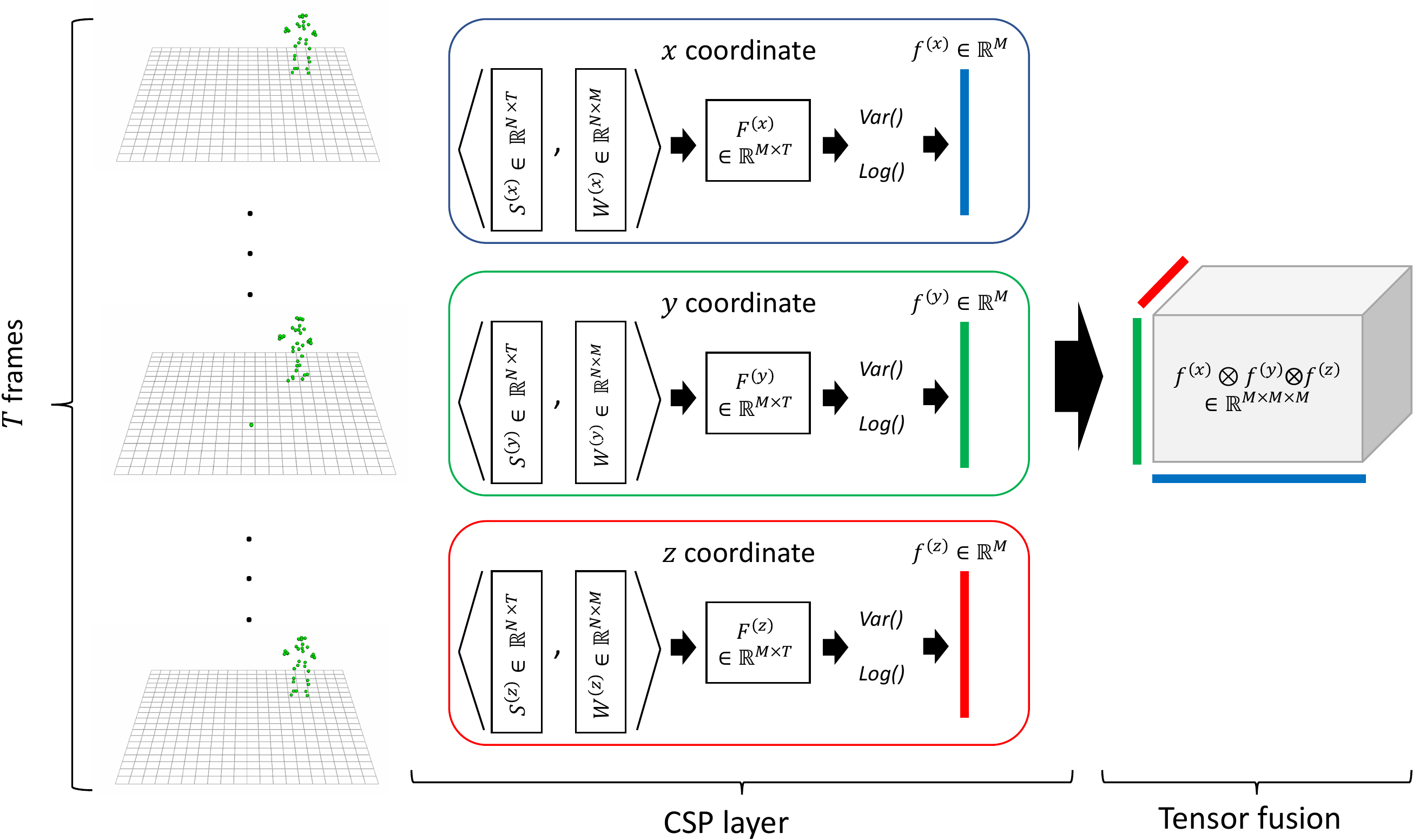}}}
	\end{minipage} 
	\caption{The proposed CSP layer and the tensor fusion operation. Parameter $N$ stands for the number of skeleton joints.}
	\label{fig:csp_fusion}
\end{figure}

\subsection{Tensor Fusion Operation}
The fusion module receives as input the feature vectors constructed by the CSP layer and produces a rich and compact representation of the data. Since we do not know in advance the kind of interactions between the elements of the constructed feature vectors, we cannot fuse them using feature averaging or addition (see Section \ref{sssec:fusion}). The employed fusion technique is motivated by the work in \cite{hu2017attribute}. The output of the fusion module corresponds to the Kronecker product of the feature vectors produced by the CSP layer. Therefore, after the fusion module each input sample $\bm S$, in the form of (\ref{eq:tensor_object}), is represented by a tensor object in $\bm X \in \mathbb R^{M \times M \times M}$. Contrary to \cite{hu2017attribute}, we do not reduce the dimensionality of the fused tensor object via decomposition techniques. Instead,  we use a tensor-based learning machine capable of processing the fused information in its original multilinear form. The proposed CSP layer and the tensor fusion operation are depicted in Fig. \ref{fig:csp_fusion}.

\subsection{Tensor-based Neural Network}
The employed tensor-based neural network is a fully connected feed forward neural network, its parameter space, however, is compressed \cite{li2018tucker}. At each layer the weights should satisfy the Tucker decomposition \cite{kolda2009tensor}. In particular, the weights $\bm W_k$ at the $k$-th hidden layer are expressed as 
\begin{equation}
\label{eq:decomposition}
\bm W_k = \bm I_k \times_1 W_k^{(1)} \times_2 W_k^{(2)} \cdots  \times_J W_k^{(J)},
\end{equation}
where $\bm I_k$ is a tensor all elements of which equal one, and the operation "$\times_j$" stands for the mode-$j$ product. 

The information is propagated through the layers of the tensor-based neural network in a sequence of projections -- at each layer the tensor input is projected to another tensor space -- and nonlinear transformations. Formally, consider a network with $(K-1)$ hidden layers. An input (tensor) sample $\bm X \in \mathbb R^{P_1 \times \cdots \times P_J}$ is propagated from the $k$-th layer of the network to the next one via the projection
\begin{equation}
\bm Z_{k+1} = \bm H_k \times_1 (W_{k+1}^{(1)})^\top \cdots \times_J (W_{k+1}^{(J)})^\top
\end{equation}
and the nonlinear transformation
\begin{equation}
\bm H_{k+1} = g(\bm Z_{k+1}),
\end{equation}
where $g(\cdot)$ is a nonlinear function (e.g. sigmoid) that is applied element-wise on a tensor object. For the input layer $\bm H_0 \equiv \bm X$. The layers that propagate information in the way described above are referred as Tensor Contraction Layers (TCL) \cite{kossaifi2017tensor}. %\footnote{The term "contraction", however, is misleading, since it implies that the projection operation should reduce the dimension of the input, which obviously is not necessary}. 

\begin{figure}[t]
	\begin{minipage}{\linewidth}
		\centering
		\centerline{\fbox{\includegraphics[width=1.0\linewidth]{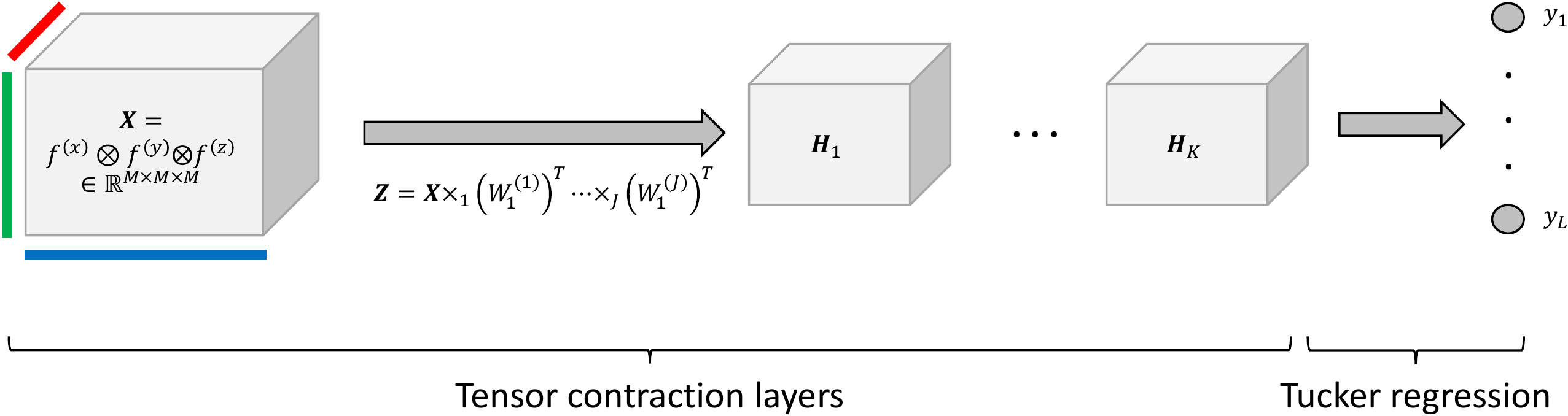}}}
	\end{minipage} 
	\caption{Propagation of information through the layers of the tensor-based neural network.}
	\label{fig:tensor_network}
\end{figure}

Finally, the output of the $(K-1)$-th hidden layer is fed to a Tucker regression model \cite{li2018tucker}, which outputs 
\begin{equation}
\label{eq:output}
y_l = s\Big( \big\langle \bm H_{K-1}, (\bm G_l \times_1 W_{K,l}^{(1)}) \cdots \times_J  W_{K,l}^{(1J)}   \big \rangle + b_l \Big)
\end{equation}
for the $l$-th class. In (\ref{eq:output}) the tensor $\bm G_l \in \mathbb R^{R_1 \times \cdots \times R_J}$ and $R_j$ is the rank of the Tucker decomposition along mode $j$ used in the output layer. The scalar $b_l$ is the bias associated with the $l$-th class, while the subscript $l$ indicates that separate sets of parameters are used to model the response for each class. The tensor-based neural network is presented inf Fig. \ref{fig:tensor_network}.

At this point it should be highlighted that the sequential projections and nonlinear transformations can be seen as a \textit{hierarchical feature construction} process, which aims to capture statistical relations between the elements of the input in order to emphasize discriminative features for the pattern recognition task. Finally, since the weights of the employed tensor-based neural network need to satisfy the decomposition in (\ref{eq:decomposition}), the total number of trainable parameters is reduced substantially \cite{li2018tucker}. This reduction acts as a very strong regularizer that shields the network against overfitting (see \cite{cichocki2017tensor}, Section 2.2).

\section{Experimental Results} 
In this section we describe the employed dataset, the effect of different parameters on the performance of the proposed scheme, and a performance evaluation of the proposed methodology against state of the art choreography modeling methods.

\subsection{Dataset Description}
\begin{figure}[t]
	\begin{minipage}{\linewidth}
		\centering
		\centerline{\fbox{\includegraphics[width=1.0\linewidth]{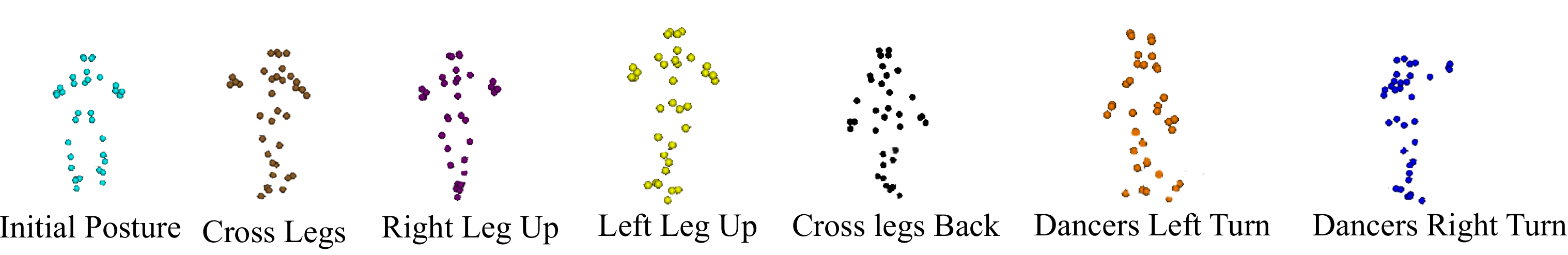}}}
	\end{minipage} 
	\caption{Examples of the seven different postures.}
	\label{fig:postures}
\end{figure}

In this study we employ the dataset captured during the framework of the EU project TERPSICHORE \cite{doulamis2017modelling}. The dataset consists of four Greek folklore dances performed by three professionals and is public available upon request. Each dance performance is described by consecutive frames and each frame is represented by the spatial coordinates of the twenty-five tracked skeleton joints (see Fig. \ref{fig:kinect}). The frames of the captured choreographies were manually annotated by dance experts according to the posture they depict. In total seven different postures are depicted, see Fig. \ref{fig:postures}. The distribution of annotated samples between different classes for each dance (performer) is depicted in Table \ref{table:dataset}, and apparently the dataset is highly unbalanced.
%Three steps are followed to transform the captured data into a dataset suitable for training and testing our proposed methodology. 
First, we follow the procedure described in Section \ref{sec:preprocessing} to transform the coordinates of skeleton joints to a coordinate system in which the origin is the Spine Base joint. Second, we use different values for parameter $T$ to create a dataset as in (\ref{eq:tensor_object}). Third, we assign to each sample the annotation of the centered frame, e.g., for $T=15$ we assign to the sample the annotation of the $8$-th frame. %By following those steps the resulting dataset consists of $\sim4000$ annotated samples. 

\begin{table}[!t]
	\centering
	\caption{Distribution of annotated samples between classes for each dance (performer).}
	\newcolumntype{L}[1]{>{\hsize=#1\hsize\raggedright\arraybackslash}X}%
	\newcolumntype{C}[1]{>{\hsize=#1\hsize\centering\arraybackslash}X}%
	\label{table:dataset}
	\begin{tabularx}{1.0\linewidth}{C{3.5}C{2.4}C{2.4}C{2.4}C{2.4}C{2.4}C{2.4}C{2.4}}
		\hline\hline
		ID & C1 & C2 &	C3 & C4 & C5 & C6 &  C7\\
		\hline\hline
		D1 (P1)   & 155 & 201 & - & - & 44 & 13 & -\\
		D2 (P1)   & 82 & 95 & 42 & 22 & - & - & 47\\
		D3 (P1)   & 122 & 246 & - & - & - & - & 82\\
		D3 (P2)   & 44 & 268 & - & - & - & - & 61\\
		D1 (P2)   & 82 & 155 & - & - & 40 & 85 & -\\
		D2 (P2)   & 82 & 112 & 16 & 32 & - & - & 44\\
		D5 (P3)   & 37 & 98 & 38 & 25 & - & - & 77\\
		D1 (P3)   & 152 & 96 & - & - & 13 & 16 & -\\
		D2 (P3)   & 33 & 102 & 38 & 25 & - & - & 77\\
		D3 (P3)   & 119 & 130 & - & - & - & - & 49\\
		\hline
		\textbf{Total}   & \textbf{908} & \textbf{1503} & \textbf{134} & \textbf{104} & \textbf{97} & \textbf{114} & \textbf{437}\\
		\hline
	\end{tabularx}
\end{table}

For evaluating the performance of our methodology, we randomly shuffle the constructed dataset and follow a $10$-fold cross validation scheme. Under that scheme, the performance is evaluated in terms of average classification accuracy and F1 score across the $10$ folds. To train our model we used Adam optimizer with learning rate equal to $2.5 \cdot 10^{-4}$. We set the maximum number of training epochs to 300 and employed early stopping criteria to avoid overfitting, which are activated if the accuracy on the validation set is not improved after 20 epochs. The validation set corresponds to 10\% of the training set for each fold. Finally, since the problem is unbalanced, we used the weighted cross entropy as the loss function, and the weight for each class corresponds to the inverse of its frequency in the training set.

\subsection{Parameters Effect Investigation}
There are three different parameters that affect the performance of the proposed methodology; namely, parameter $M$, that is the dimension of feature vector constructed by the CSP layer, parameter $T$, that is the temporal dimension of the samples, and $K$ that is the number of tensor contraction layers employed in the tensor-based neural network architecture.

\subsubsection{The effect of parameter $M$}
Parameter $M$ corresponds to the dimension of the features constructed by the CSP layer.% of the proposed neural network architecture. 
For investigating the effect of that parameter on the performance of the model, we keep fixed the parameter $T=7$. Then, we train and test the performance of the proposed model with two tensor contraction layers (TCLs) for different values of $M$, i.e., $M=12$, $M=18$, $M=24$, and $M=30$. The dimension of the tensor contraction and regression layers is presented in the second column of Table \ref{table:projections}.% receives an input in $\mathbb R^{M \times M \times M}$ and projects it in $\mathbb R^{4 \times 4 \times 4}$ and the ranks of the Tensor Regression Layer are $2 \times 2 \times 2$.

\begin{figure}[t]
	\begin{minipage}{\linewidth}
		\centering
		\centerline{\fbox{\includegraphics[width=1.0\linewidth]{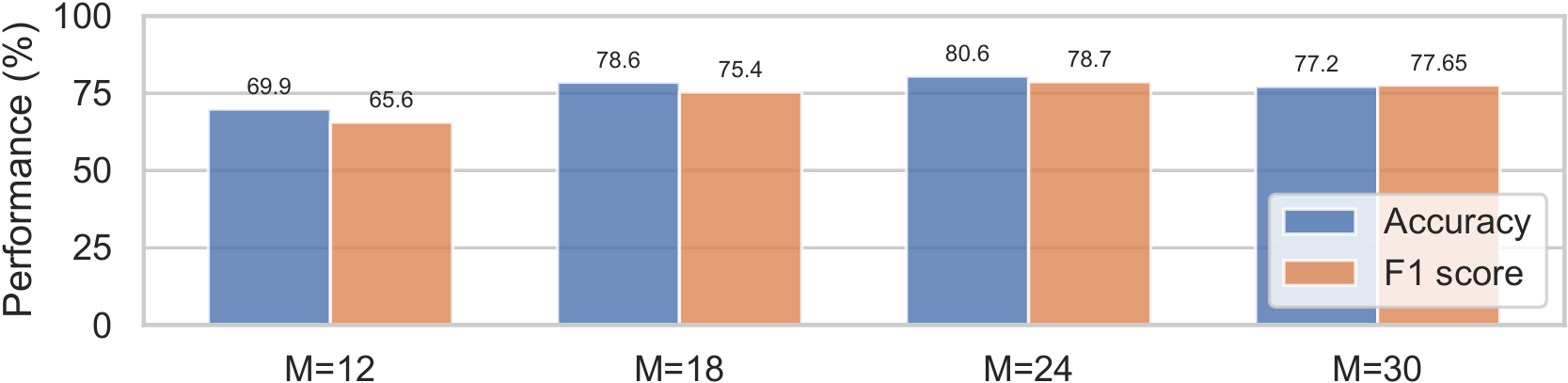}}}
	\end{minipage} 
	\caption{Average classification accuracy and F1 score of a tensor-based neural network with two TCLs, for $T=7$, and for different values of $M$.}
	\label{fig:parameterM}
\end{figure}

The effect of the parameter $M$ is depicted in Fig. \ref{fig:parameterM}. %By increasing the value of parameter $M$ the performance of the proposed model monotonically increases. 
The best accuracy is achieved for $M=24$. The dimension of the features constructed by the CSP layer is directly related to their representation power. Thus, features of higher dimension can better capture the spatial and temporal patterns of skeleton data resulting to more accurate human pose classification. For $M=30$, however, the accuracy drops, which might be an indication of over-fitting. Moreover, increasing the value of parameter $M$ increases the total number of trainable parameters of the model. %, and thus, the learning capacity of the model. This further justifies higher performance for larger values of $M$. 
Indicatively, the number of trainable parameters for $M$ equals $12$, $18$, $24$ and $30$ is $1335$, $1839$, $2343$, and $2847$ respectively.

\subsubsection{The effect of parameter $T$}
\label{sssec:parameterT}
In contrast to parameter $M$, parameter $T$ does not affect the number of trainable parameters of the model nor the dimension of the features constructed by the CSP layer due to the variance operator employed in (\ref{eq:csp_features}). Parameter $T$ indirectly determines the amount of temporal information that is taken into consideration during the construction of the features. %Therefore, small values of $T$ result to features that encode small amount of temporal information and may not be able to sufficiently represent the temporal relations present in the data. 

\begin{figure}[t]
	\begin{minipage}{\linewidth}
		\centering
		\centerline{\fbox{\includegraphics[width=1.0\linewidth]{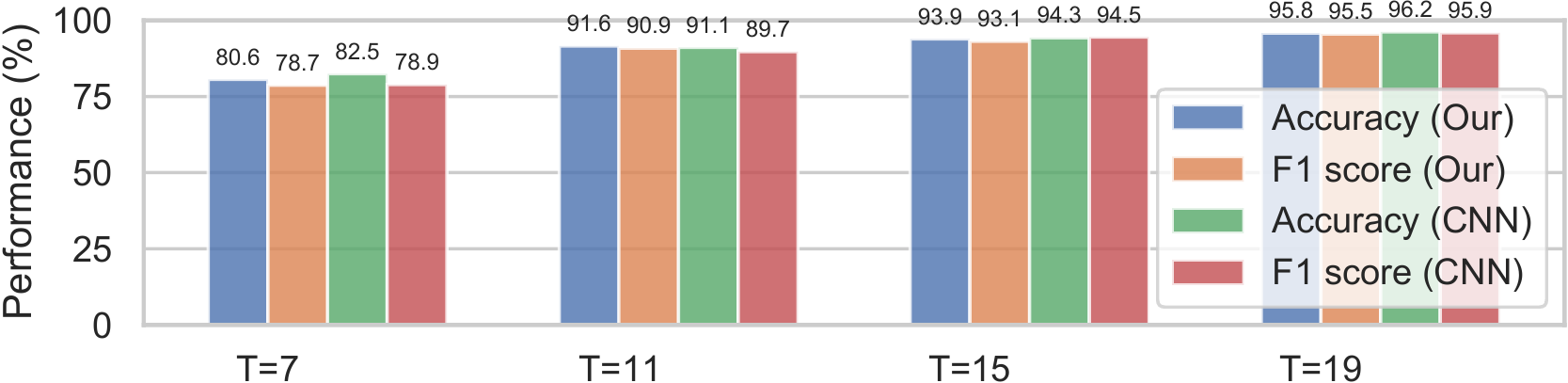}}}
	\end{minipage} 
	\caption{Average classification accuracy and F1 score of a tensor-based neural network with two TCLs, for $M=24$, and for different values of $T$.}
	\label{fig:parameterT}
\end{figure}

The effect of parameter $T$ on the performance of the model is presented in Fig. \ref{fig:parameterT}. To obtain those results we train a tensor-based neural network with two TCLs and keep the value of parameter $M$ fixed equal to $24$. %Again, the TCLs receives an input in $\mathbb R^{24 \times 24 \times 24}$ and projects it in $\mathbb R^{4 \times 4 \times 4}$ and the ranks of the Tensor Regression Layer are $2 \times 2 \times 2$.
Producing features that encode larger amounts of temporal information results to higher human pose recognition accuracy. Increasing the value of parameter $T$  from $7$  to $11$ results in a performance improvement more than $10\%$. Increasing, however, more the value of $T$ results in smaller performance improvements around $2\%$. This implies that capturing important temporal information for problem at hand more that $11$ consecutive frames need to be used. 

In Fig. \ref{fig:parameterT} we also compare the performance of the proposed model against a 1D-CNN. First, we concatenated the measurements of different channels to produce input samples for the CNN of dimension $72 \times T$. The CNN performs convolutions along the temporal dimension of the samples, and thus, similarly to the proposed model, it encodes the temporal information within the constructed feature vectors. The employed CNN consists of 3 convolutional layers with 8, 16 and 24 kernels, which are followed by a dense layer with 12 neurons and the output layer. The width of the kernels is $(T-1)/2$ for the first two layers and 3 for the third layer. The 1D-CNN and the proposed model perform almost the same. The proposed model, however, employs a significantly smaller number of trainable parameters. Specifically, the proposed model employs $2343$ trainable parameters, while the CNN employs $3415$, $4631$, $5847$ and $7063$ trainable parameters for $T=7, 11, 15$ and $19$ respectively.   

\subsubsection{The effect of parameter $K$}
\begin{figure}[t]
	\begin{minipage}{\linewidth}
		\centering
		\centerline{\fbox{\includegraphics[width=1.0\linewidth]{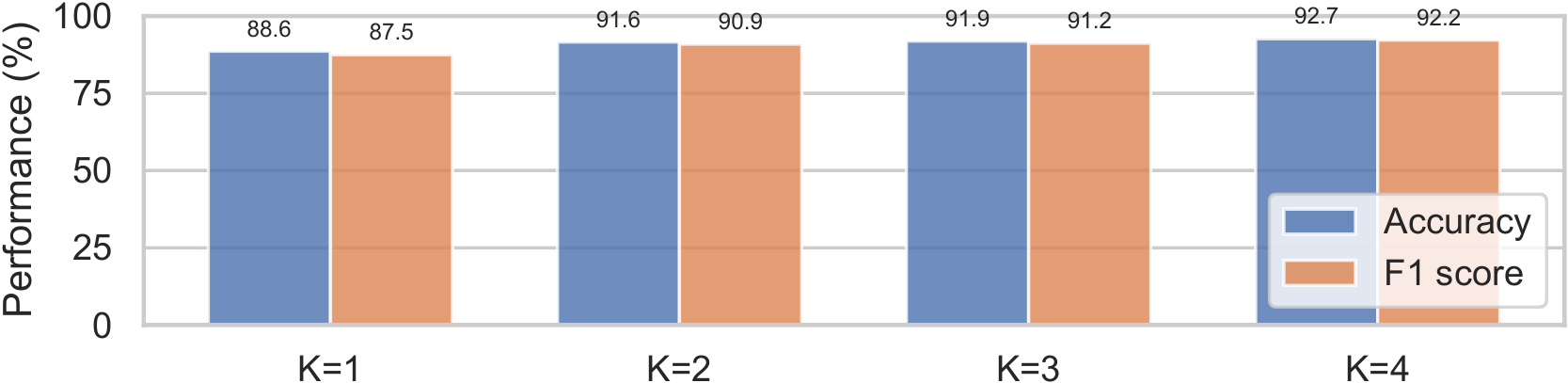}}}
	\end{minipage} 
	\caption{Average classification accuracy and F1 score of a tensor-based neural network with different number of  tensor contraction layers (parameter $K$) for $M=24$ and $T=11$.}
	\label{fig:parameterK}
\end{figure}

Parameter $K$ corresponds to the number of TCLs present in the network. Fig. \ref{fig:parameterK} presents the effect of the number of TCLs on the performance of the model. To obtain those results we keep parameter $M$ an $T$ fixed and equal to $24$ and $11$ respectively, and trained four different tensor-based neural networks with $1$, $2$, $3$, and $4$ tensor contraction layers. The projections of the employed contraction layers are presented in Table \ref{table:projections}. Increasing the number of tensor contraction layers increases the total number of trainable parameters of the model, and thus its learning capacity. Indicatively, the number of trainable parameters for $K$ equals $1$, $2$, $3$, and $4$ is $1959$, $2343$, $2919$, and $3783$ respectively. That increase, however, does not seem to affect the performance of the model, since the performance improvement from $K=2$ to $K=4$ is only $1\%$.

\begin{table}[!t]
	\centering
	\caption{Projections of tensor objects when they propagated through Tensor Contraction Layers (TCL) and the ranks of the Tensor Regression Layer (TRL).}
	\newcolumntype{L}[1]{>{\hsize=#1\hsize\raggedright\arraybackslash}X}%
	\newcolumntype{C}[1]{>{\hsize=#1\hsize\centering\arraybackslash}X}%
	\label{table:projections}
	\begin{tabularx}{1.0\linewidth}{L{2}C{5.7}C{5.7}C{5.7}C{5.7}}
		\hline\hline
		& 1 TCL &2 TCLs & 3 TCLs &	4 TCLs  \\
		\hline\hline
		Input   & $(24 \times 24 \times 24)$ & $(24 \times 24 \times 24)$ &$(24 \times 24 \times 24)$ &$(24 \times 24 \times 24)$  \\
		Layer1& $(4 \times 4 \times 4)$ & $(8 \times 8 \times 8)$ &  $(12 \times 12 \times 12)$ & $(16 \times 16 \times 16)$\\
		Layer2& - & $(4 \times 4 \times 4)$ &  $(8 \times 8 \times 8)$ & $(12 \times 12 \times 12)$\\
		Layer3& - & - &  $(4 \times 4 \times 4)$ & $(8 \times 8 \times 8)$\\
		Layer4& - & - & - & $(4 \times 4 \times 4)$\\
		TRL& $(2 \times 2 \times 2)$ & $(2 \times 2 \times 2)$ & $(2 \times 2 \times 2)$ & $(2 \times 2 \times 2)$\\
		\hline
	\end{tabularx}
\end{table}

The investigation above suggests that the most important parameter for achieving highly accurate results is parameter $M$. Indeed, increasing the dimension of the features constructed by the CSP layer from $12$ to $24$, we achieve a performance improvement of more than $10\%$. On the contrary, designed deeper architectures does not seem to significantly affect the performance of the model.
This might be due to the Tucker decomposition % imposed on the weights of the tensor contraction layers 
(see (\ref{eq:decomposition})), which acts as a very strong regularizer for the model.

\subsection{Performance Evaluation Against State of the Art Methods}
In this section we compare the performance of the proposed model against state-of-the-art methods for choreographic modeling. We compare the performance of our model against LSTM and the recently proposed Bayesian Optimized Bidirectional LSTM (BOBi LSTM) \cite{rallis2019learning}.
In contrast to the proposed model and the 1D-CNN, the LSTM-based models exploit the order of the data as an additional source of information.

For the performance comparison, we utilize a tensor-based neural network with two TCLs ($K=2$), and parameters $M$ and $T$ equal to $24$ and $11$ respectively. Regarding the LSTM and the BOBi LSTM models, their architectures are the ones presented in \cite{rallis2019learning} and they use a memory of $10$ frames for recognizing human poses. At this point we should emphasize that those models receive as input the kinematic properties of the skeleton joints; i.e., the spatial position as well as the velocity and the acceleration of each joint. In contrast, our method receives as input \textit{solely} the spatial position of the joints. Moreover, the proposed model consists of $2343$ trainable parameters. In contrast, the BOBi-LSTM network in \cite{rallis2019learning} was composed by 2 LSTM Layers of 128 cells each and two additional dense layers as the output. This makes the total number of training parameters at 205,674, namely \textit{87 times} more than the number of trainable parameters in our approach. This significant reduction favors the efficient parameter estimation especially when small sample setting problems need to be addressed.

Table \ref{table:SoA} presents the results of that comparison. The proposed model performs more than $6\%$ better compared the BOBi LSTM, despite the fact that is uses a simpler input representation (our method is completely blind to kinematics information of the skeleton joints). Also, Table \ref{table:SoA} presents the performance of the 1D-CNN mentioned above. The 1D-CNN performs better than both LSTM models and slightly worse than our proposed model. This implies that models that do not take into consideration the order of the samples are more appropriate for classifying human poses in folklore dances. This is justified by the fact that different dances are composed of different sequences of poses. Therefore, information regarding the order of the samples confuses the model and deteriorates its performance.

\begin{figure}[t]
	\begin{minipage}{\linewidth}
		\centering
		\centerline{\fbox{\includegraphics[width=1.0\linewidth]{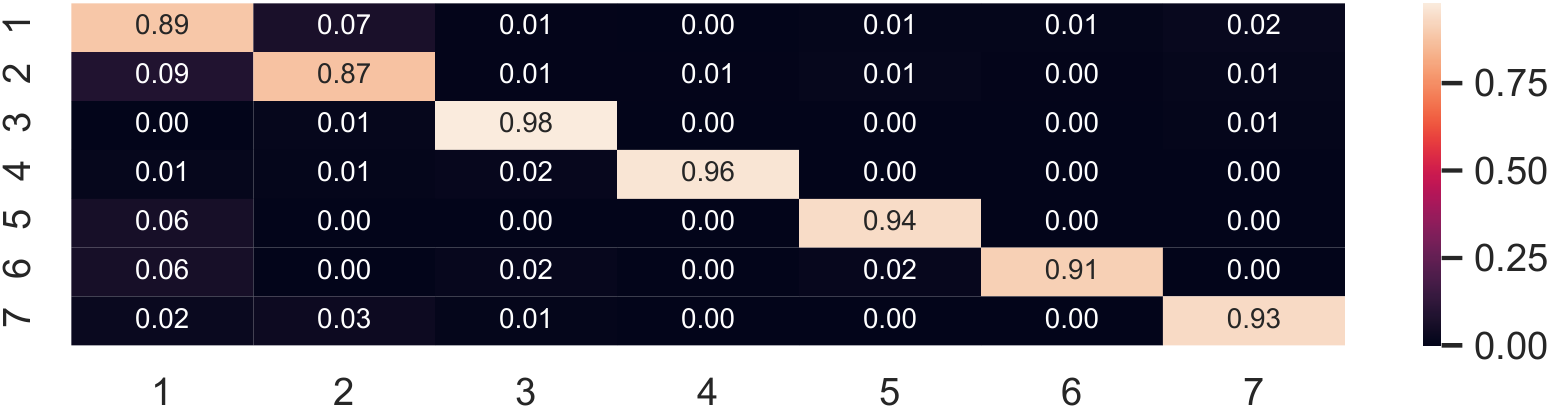}}}
	\end{minipage} 
	\caption{Confusion matrix for $T=11$, and $M=24$.}
	\label{fig:confusion_matrix}
\end{figure}

Fig. \ref{fig:confusion_matrix} presents the confusion matrix for the proposed model. The models performs very well for all classes with the smallest accuracy to be 87\% for the second class (cross-legs). 9\% of the samples belonging to the second class are misclassified to class 1 (initial pose). This mainly happens due to similarities of the poses belonging to these two classes. For poses that belong to the first and the second classes the dancer faces the camera, and the measurements for all joints (except knees and ankles) are very similar.

%Most of the misclassified samples come from class 5 (cross-leg back); 43\% of the samples belonging to it are classified in class 1 (initial pose). This may happen for two reasons. First, the 5-th class is the most under-represented class in the dataset. Second, there are similarities between these two classes, since for both the dancer faces the camera, and the measurements for all joints (except knees and ankles) are very similar.}

The comparison above implies the following. First, the proposed CSP layers can produce highly discriminative features that encode the spatial and the temporal information in the data. Second, employing the tensor fusion operation produces compact yet highly descriptive representations of the input. Finally, tensor contraction and tensor regression layers can efficiently process data in tensor form and produce hghly accurate learning models.

\begin{table}[!t]
	\centering
	\caption{Performnce comparison in terms of average classification accuracy and F1 score against LSTM and BOBi LSTM models.}
	\newcolumntype{L}[1]{>{\hsize=#1\hsize\raggedright\arraybackslash}X}%
	\newcolumntype{C}[1]{>{\hsize=#1\hsize\centering\arraybackslash}X}%
	\label{table:SoA}
	\begin{tabularx}{0.98\linewidth}{L{7}C{10}C{10}}
		\hline\hline
		 &Accuracy (\%) & F1 Score (\%)	\\
		\hline\hline
		LSTM              & 	84.2\%    &82.0\%  \\
		BOBi LSTM     &    85.4\%     &80.7\% \\
		1D-CNN    &    91.1\%     &89.7\% \\
		Our Approach &     \textbf{91.6\%}     &\textbf{90.9\%}\\
		\hline
	\end{tabularx}
\end{table}

\section{Conclusion}
In this study we proposed a spatially and temporally aware tensor-based neural network that can efficiently process spatiotemporal data. We evaluated the performance of the proposed model on the problem of human pose classification using 3D data captured using the Kinect-II sensor. The evaluation results indicate that the proposed model can construct highly discriminative spatiotemporal features and achieve state-of-the-art performance. The evaluation of the proposed model was carried out following a 10 fold cross validation protocol. Following a leave-one-dancer-out protocol is more appropriate for evaluating the generalization ability of learning models. For this dataset, however, the number of dancers (2 males – 1 female) is not adequate to train/evaluate the model using that very demanding protocol. Indeed, the performance of the proposed model and the 1D-CNN when trained and evaluated using the leave-one-subject-out scheme is $\sim52\%$, that is 8\% better than a dummy classifier that always predicts the majority class of the training set. One way to improve the performance under this highly demanding protocol, is to employ application specific data transformations, such as warping techniques for view invariance, and normalization methods to cope with dancers of different heights. Investigation of such techniques is the priority of our future work.

%To conclude, as mentioned in Section\ref{ssec:formulation}, the problem of recognizing human poses using 3D skeleton data is a specific instance of the more general problem of pattern recognition using information coming from sensor network. Therefore, despite the fact that in this study we consider that specific problem, our model is a general one that can potentially be applied on general pattern recognition problems that employ spatiotemporal data from sensor networks. 

\section*{Acknowledgement}
This work has been supported by the European Union's Horizon 2020 research and innovation programme from the TAMED project (Grant Agreement No. 101003397).


%\section*{Acknowledgment}




%\bibliographystyle{IEEEtran}
%\bibliography{IEEEabrv,refs}
%\begin{thebibliography}{1}
%\end{thebibliography}

\end{document}